\documentclass[conference]{IEEEtran}
\IEEEoverridecommandlockouts
\usepackage{cite}
\usepackage{amsmath,amssymb,amsfonts}
\usepackage[linesnumbered,ruled,vlined]{algorithm2e}
\SetKwInput{KwInput}{Input}                
\SetKwInput{KwOutput}{Output}              
\usepackage{graphicx}
\usepackage[export]{adjustbox} 
\usepackage{textcomp}
\usepackage{xcolor}
\def\BibTeX{{\rm B\kern-.05em{\sc i\kern-.025em b}\kern-.08em
    T\kern-.1667em\lower.7ex\hbox{E}\kern-.125emX}}

\usepackage{hyperref}
\hypersetup{
     colorlinks = true,
     linkcolor = red,
     anchorcolor = black,
     citecolor = green,
     filecolor = cyan,
     menucolor = red,
     runcolor = cyan, 
     urlcolor = magenta,
     }

\usepackage[nameinlink,noabbrev]{cleveref}

\Crefname{equation}{}{Eqs.}         
\Crefname{figure}{Fig.}{Figs.}

\usepackage{tikz}

\usetikzlibrary{arrows}
\usetikzlibrary{positioning, shapes.geometric, shapes.misc}

\usepackage{multirow}
\usepackage{siunitx}
\usepackage{bbm}
\usepackage[normalem]{ulem}

\usepackage{graphicx,subcaption,lipsum}

\newcommand{\mIoU}{\mathit{mIoU}}

\usepackage{enumitem}

\begin{document}


\title{A Preprocessing and Postprocessing Voxel-based Method for LiDAR Semantic Segmentation Improvement in Long Distance
}

\author{

\IEEEauthorblockN{1\textsuperscript{st} Andrea Matteazzi\textsuperscript{1,2}}
\IEEEauthorblockA{
matteazzi@uni-wuppertal.de}

\and 

\IEEEauthorblockN{2\textsuperscript{nd} Pascal Colling\textsuperscript{2}}
\IEEEauthorblockA{
pascal.colling@aptiv.com}

\and 

\IEEEauthorblockN{3\textsuperscript{rd} Michael Arnold\textsuperscript{2}}
\IEEEauthorblockA{
michael.arnold@aptiv.com}

\and 

\IEEEauthorblockN{4\textsuperscript{th} Dietmar Tutsch\textsuperscript{1}}
\IEEEauthorblockA{
tutsch@uni-wuppertal.de}

\and
\centerline{\textsuperscript{1}\textit{University of Wuppertal}}
\and
\centerline{\textsuperscript{2}\textit{Aptiv Services Deutschland GmbH}}

}

\maketitle
 
\begin{abstract}
In recent years considerable research in LiDAR semantic segmentation was conducted, introducing several new state of the art models. However, most research focuses on single-scan point clouds, limiting performance especially in long distance outdoor scenarios, by omitting time-sequential information. Moreover, varying-density and occlusions constitute significant challenges in single-scan approaches. In this paper we propose a LiDAR point cloud preprocessing and postprocessing method. This multi-stage approach, in conjunction with state of the art models in a multi-scan setting, aims to solve those challenges. We demonstrate the benefits of our method through quantitative evaluation with the given models in single-scan settings. In particular, we achieve significant improvements in $\mIoU$ performance of over $5$ percentage point in medium range and over $10$ percentage point in far range. This is essential for 3D semantic scene understanding in long distance as well as for applications where offline processing is permissible.
\end{abstract}


\section{Introduction} \label{sec:introduction}

\begin{figure}[h]
\centering
  \begin{subfigure}{0.40\textwidth}
  \setlength{\fboxrule}{0.5pt}
    \framebox{\includegraphics[width=.9\linewidth]{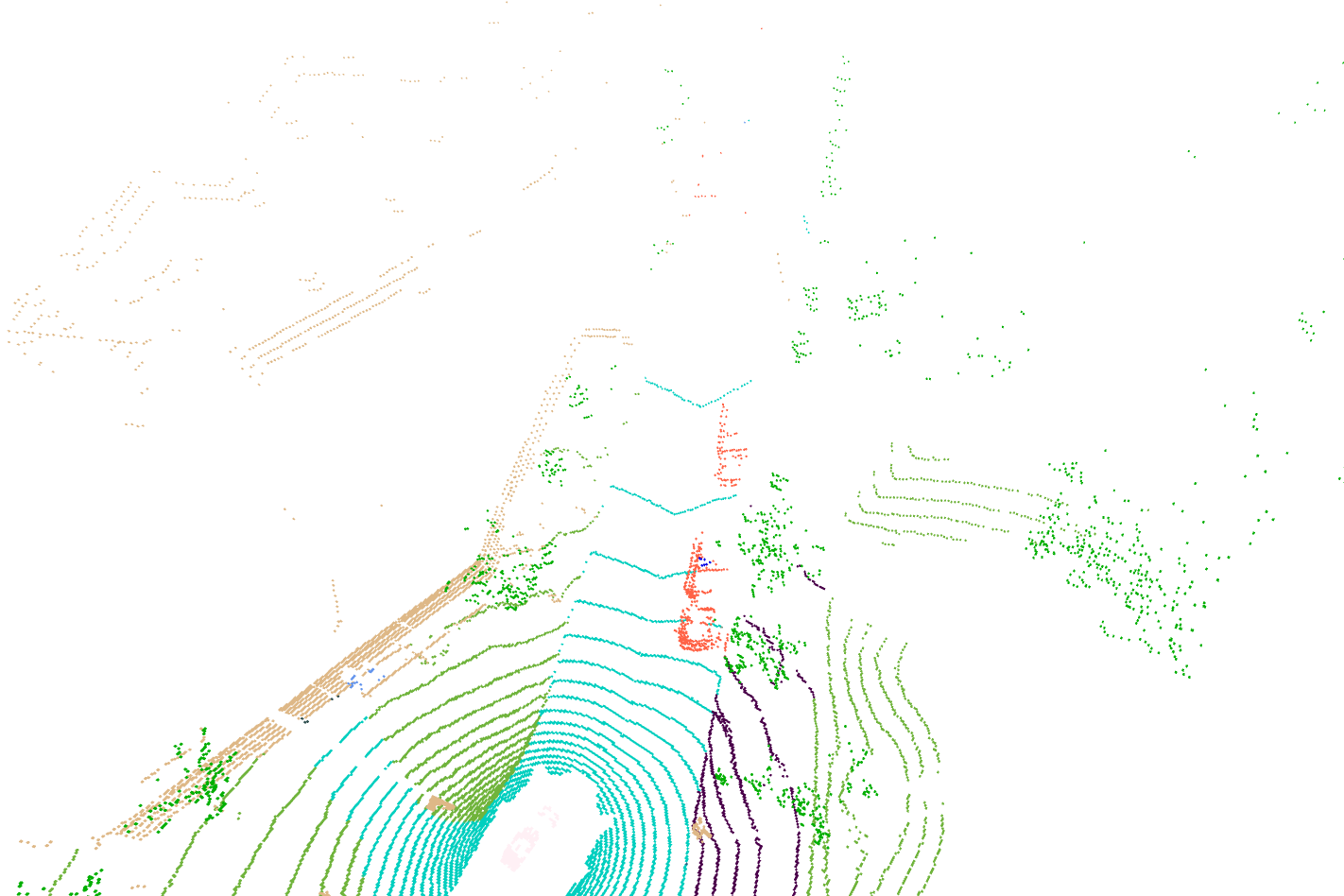}}
    \caption{single-scan}
  \end{subfigure}%
  \\
  \begin{subfigure}{0.40\textwidth}
  \setlength{\fboxrule}{0.5pt}
    \framebox{\includegraphics[width=.9\linewidth]{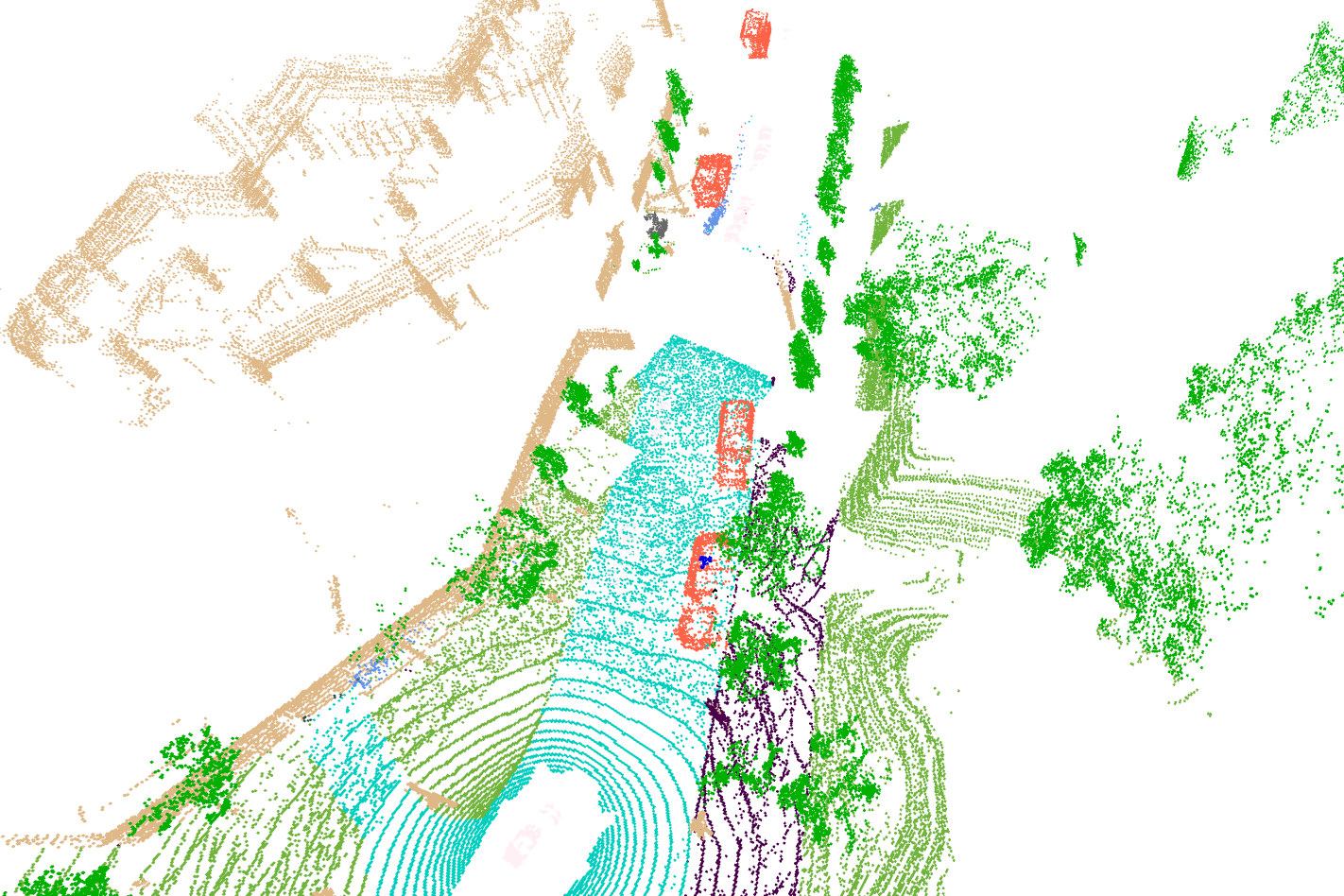}}
    \caption{multi-scan}
  \end{subfigure}%
\caption{Typical voxel distribution for LiDAR outdoor point cloud scenarios on nuScenes (a) and the consistent voxel distribution achieved by our preprocessing method (b). It is worth to mention that the distribution in (b) is obtained with a limited amount of voxels, based on our novel downsampling technique.}
\label{fig:single_multi_1}
\end{figure}

LiDAR sensor is a well known technology used for autonomous driving perception tasks. Indeed, it can provide very detailed 3D mapping of the surrounding environment, with higher resolution and better accuracy compared to other sensors. For example, it provides better depth information compared to camera and higher resolution data compared to radar.
Semantic segmentation of LiDAR point cloud data is a fundamental research topic and several manual annotated datasets were created for such a task. In this paper, we focus on two important datasets, namely SemanticKITTI \cite{behley2019semantickitti} and nuScenes \cite{caesar2020nuscenes}.
Since in recent years many voxel-based LiDAR semantic segmentation architectures were developed \cite{zhou2020cylinder3d, zhu2020cylindrical, tang2020searching, lai2023spherical}, this paper deals with the further improvement of them in order to enhance the results. In the next section, we will motivate the choice to focus on voxel-based methods in more detail. 
In particular, this paper wants to provide a general voxel-based method to improve the semantic segmentation performance in medium and far ranges. Indeed, while most of research focuses on single-scan models \cite{tang2020searching, kong2023rethinking, puy2023using, zhou2020cylinder3d, zhu2020cylindrical, lai2023spherical}, multi-scan approaches are limited to few point clouds, aiming at detecting motion states (e.g., moving or static) of points in an online fashion\cite{liu2023mars3d, duerr2021lidarbased}. This limits the semantic segmentation performance of models in farther ranges, because the density of points becomes sparser for long distance. This is mainly due to the so called \emph{varying-density property} and additionally, to possible \emph{occlusion of objects}. In our work however, the use of multiple point clouds aims to improve the performance of models in an offline fashion. Hence, our approach exploits a considerable amount of point clouds and solves the overhead in data complexity by a novel downsampling technique. An example of applying our preprocessing method is shown in Fig. \ref{fig:single_multi_1}. Moreover, a postprocessing method further improves the performance by applying an ensemble modeling and a weighted voting scheme. Overall, the method permits to overcome the limitation of semantic segmentation quality in long distance, by increasing the density of points in farther range. This turns out to be beneficial for many offline tasks, like ground truth generation, where the correct semantic segmentation of distant points may be crucial. Another important aspect to notice is that most of the research in point cloud semantic segmentation aims to improve the overall $\mIoU$. However, since most of the points reside in the close range, improvements made by new models and approaches mainly translate in better close range semantic segmentation. Indeed, a big improvement in the semantic segmentation of farther points is not reflected in the overall $\mIoU$, because their sparsity drastically limits their contribution in the overall performance. In our work, we instead break down the $\mIoU$ in order to carefully analyze the variation of performances for different ranges, and we build our method on top of that, achieving improvements of over $5$ percentage point for medium range and over $10$ percentage point for far range.\\ 
In summary, the contribution of this paper is:
\begin{itemize}
    \item Developing a general preprocessing and postprocessing method for improving LiDAR point clouds semantic segmentation in long distance.
    \item Prove the enhancement given by the method, by deploying it on an existing voxel-based state of the art (SOTA) architecture.
\end{itemize}

\section{Related Work} \label{sec:Related}

\subsection{Outdoor Point Cloud Semantic Segmentation} \label{subsec:Outdoor}

A point cloud can be defined as a set of 3D points
$\{p_i|i=1, \ldots ,n\}$, where each point $p_i$ is a vector of its $(x_i, y_i, z_i)$ coordinate plus extra feature channels such as intensity \cite{qi2017pointnet}. Point cloud semantic segmentation is a dense-prediction task where, given a set of 3D points $P=\{p_i|i=1, \ldots ,n\}$ and a label set $Y=\{y_j|j=1, \ldots ,k\}$,
we need to assign one of the $k$ semantic labels to each input point $p_i$. Generally speaking, LiDAR point clouds are complex data with many challenges \cite{li2020deep}. Indeed, point clouds are irregular structures with a permutation invariant problem. Moreover, there are variabilities in point clouds given by e.g., rigid transformations and local, long-range and multi-scale features to take into account. Furthermore, unlike indoor point clouds \cite{dai2017scannet, Armeni_2016_CVPR} with generally uniform density and small ranges of the scenes, outdoor point clouds usually cover a very large area, with a range up to and over $100$m. Hence, they generally contain much more points but are much sparser than those of the indoor scenes. In particular, the farther the points are, the sparser they are (varying-density property) and in the specific case of driving scenarios, it is very likely to have occlusion of objects. 

\subsection{Voxel-based methods} \label{subsec:Voxel} 

Due to the mentioned challenges around outdoor point clouds, in this paper we focus on voxel-based approaches for performing semantic segmentation. Indeed, they permit to divide the 3D space occupied by the point cloud into a regular grid of small 3D cells, called voxels. This provides a fixed and regular structure to point clouds, that can be easily processed by sparse convolution \cite{choy20194d, graham20173d, graham2017submanifold}.
Cylinder3D \cite{zhou2020cylinder3d} introduces a cylindrical partition to leverage the varying-density property of LiDAR point clouds.
Hence, differently from point-based methods, which operate directly on the unordered point sets by aggregating information from neighbors \cite{qi2017pointnet, wu2022point, lai2022stratified}, they permit an efficient computation. Moreover, differently from view-based methods \cite{kong2023rethinking, wu2018squeezesegv2, xu2021squeezesegv3}, which transform the point cloud in a two-dimensional range view, they enable to maintain 3D information. Therefore, the preprocessing and postprocessing method is developed based on a voxel-based approach, to better suit the downstream voxel-based semantic segmentation model. In our work we employe SphereFormer \cite{lai2023spherical} as downstream model and in the next section we will analyze and motivate this choice.

\subsection{Multi-scan methods} \label{subsec:Multi}

Most research in multi-scan semantic segmentation focuses on detecting motion states (e.g., moving or static) of points based on multi-scan point cloud data \cite{wang2022sequential, 9156504, liu2023mars3d, schütt2022abstract}.
Consequently, the fusion of multiple point clouds to form a single point cloud is usually straightforward and bounded to few and temporally close point clouds. On the other hand, there are some early attempts to employ recurrent networks and attention modules to fuse information across different temporal frames \cite{wang2022sequential, duerr2021lidarbased, Wang_2022}. SpSequenceNet \cite{9156504} proposes a U-Net based architecture to extract frame features between only two consecutive voxelized point clouds. The features are then combined to gather temporally global and local information. TemporalLatticeNet \cite{schütt2022abstract} proposes to match similar abstract features between few adjacent frames and fuse them temporally.
However, these approaches usually do not perform well due to the insufficiency of temporal representations and the limited feature extraction ability of the model. Our approach instead is developed in order to solely improve the single-scan semantic segmentation and it does not require any specific voxel-based model architecture. The aggregation of point clouds is indeed carefully designed in order to improve the performance in long distance for any voxel-based model. Furthermore, the postprocessing stage further boosts the performance by exploiting the multi-scan semantic segmentation.

\section{Our Method} \label{sec:method}

\begin{figure}[ht]
    \centering
    \centerline{\includegraphics[width=0.55 \textwidth]{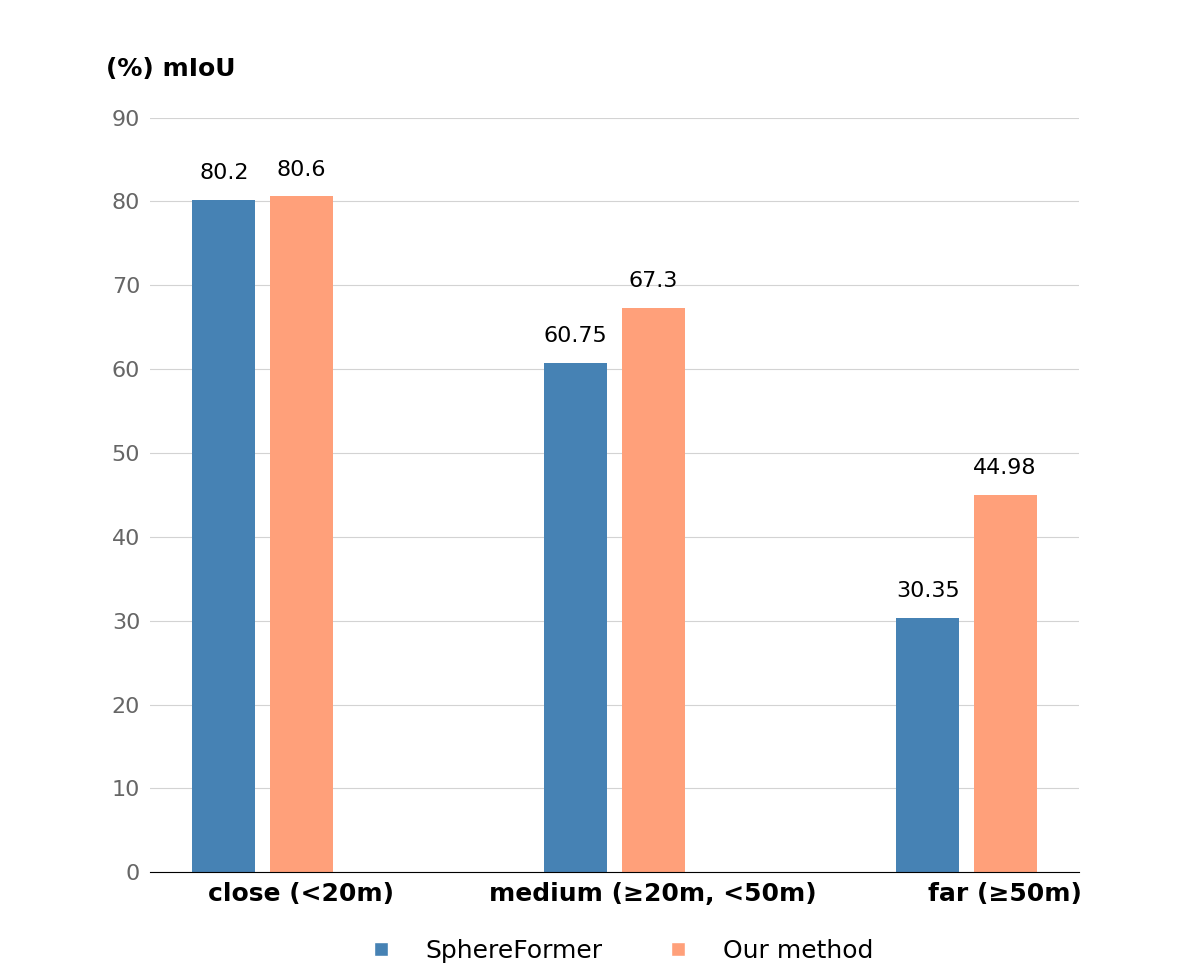}}
    \caption{Semantic segmentation performance on nuScenes val set at different ranges for SphereFormer (baseline) and our novel method. 
    }
    \label{fig:sphere_ranges}
\end{figure}

\begin{figure}[h]
\centering
  \begin{subfigure}{0.40\textwidth}
  \setlength{\fboxrule}{0.5pt}
    \framebox{\includegraphics[width=.9\linewidth]{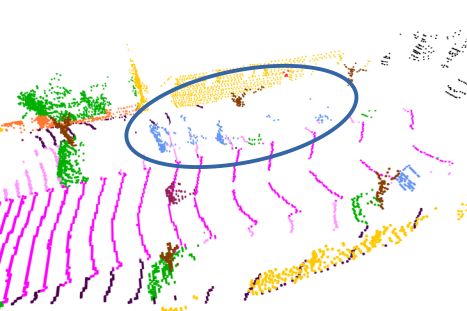}}
    \caption{single-scan}
  \end{subfigure}%
  \\
  \begin{subfigure}{0.40\textwidth}
  \setlength{\fboxrule}{0.5pt}
    \framebox{\includegraphics[width=.9\linewidth]{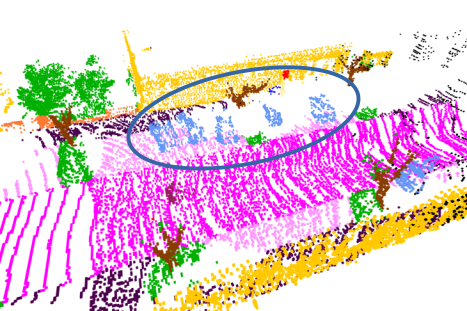}}
    \caption{multi-scan}
  \end{subfigure}%
\caption{Density of voxels representing cars (blue in highlighted area) in medium range on SemanticKITTI for single-scan (a) and for our preprocessing multi-scan (b).}
\label{fig:single_multi_2}
\end{figure}

Most of the research in LiDAR point cloud semantic segmentation is focused on improving the overall $\mIoU$ score. However $\mIoU$ score achieved for different ranges varies considerably and it decreases for farther ranges. In particular, it is possible to individuate three regions of sparsity indicated in range of meters: close ($ <20$m), medium ($\geq20$m, $<50$m) and far ($\geq 50$m).
Those intervals are chosen accordingly with the analysis of SphereFormer \cite{lai2023spherical}, in order to have a fair comparison.
Hence, it is possible to correlate the performance of a semantic segmentation model, in terms of $\mIoU$, with the sparsity of the region itself, Fig. \ref{fig:sphere_ranges}.
This is due to the varying-density property of point clouds. Indeed, the farther the points are with respect to the ego-vehicle, the sparser they are and hence, the more difficult is their segmentation, Fig. \ref{fig:single_multi_1}.
In fact, underlying objects may be underrepresented by sparser points like in the case of cars in Fig. \ref{fig:single_multi_2}.

\subsection{Preprocessing} \label{subsec:preprocessing}

A typical LiDAR point cloud dataset is made of several sequences. Each sequence corresponds to a recording of the surrounding environment for a certain period of time. Hence, it is made up of several point clouds, each of them corresponding to a timestamp of such recording.
Since this approach is offline, we can relax the restriction of compute time and demand but there can be limitations due to, for example, to the maximum number of voxels allowed by the specific downstream semantic segmentation model.
The idea of the preprocessing is then to exploit each sequence. In particular, for each timestamp of the given sequence, it accumulates point clouds in the past and in the future with respect to the considered timestamp and associated point cloud (\emph{reference point cloud}). Specifically, the accumulation is performed for the medium and far ranges, to make these regions denser. This allows to provide a better context to sparse distant points, as in the case of cars in Fig. \ref{fig:single_multi_2}. Consequently, it permits to shift semantic segmentation performance of farther ranges towards the performance of the close range.\\
The preprocessing method consists of four steps and, starting from a sequence of point clouds, it generates a sequence of multi-scan point clouds with a limited amount of consistently dense voxels.

\paragraph{A distant-based algorithm to select windows of point clouds to be accumulated, one for each timestamp}
In order to have an efficient point cloud accumulation, only a fixed amount  \texttt{accumulate\_length} of point clouds can be selected for each timestamp.
Furthermore, in each sequence, the ego-vehicle is moving in an environment with a non-constant velocity, that can be described through the translation of the vehicle $t=(t_{x}, t_{y}, t_{z})$. Hence, not always the temporal closest point clouds, with respect to the reference point cloud, can guarantee a meaningful accumulation. For example, if the ego-vehicle is not moving for several consecutive timestamps, the accumulation of those corresponding point clouds would not bring any extra context to the reference points.
For these reasons, a distant-based algorithm is developed. In particular, every two consecutive point clouds selected need to have a distance, in terms of $L_2$ norm of the translation, greater or equal than a threshold \texttt{min\_dist} meters. The resulting window of point clouds is given by the \texttt{accumulate\_length} closest point clouds to the reference point cloud, with at least \texttt{min\_dist} meters distance between two consecutive ones. A reasonable choice of these two parameters can permit to be robust to an high variance of velocity, as it allows to accumulate point clouds covering an overall range of around $\texttt{accumulate\_length} * \texttt{min\_dist}$ meters.

\begin{figure}[h]
\centering
  \begin{subfigure}{0.40\textwidth}
  \setlength{\fboxrule}{0.5pt}
    \framebox{\includegraphics[width=.9\linewidth]{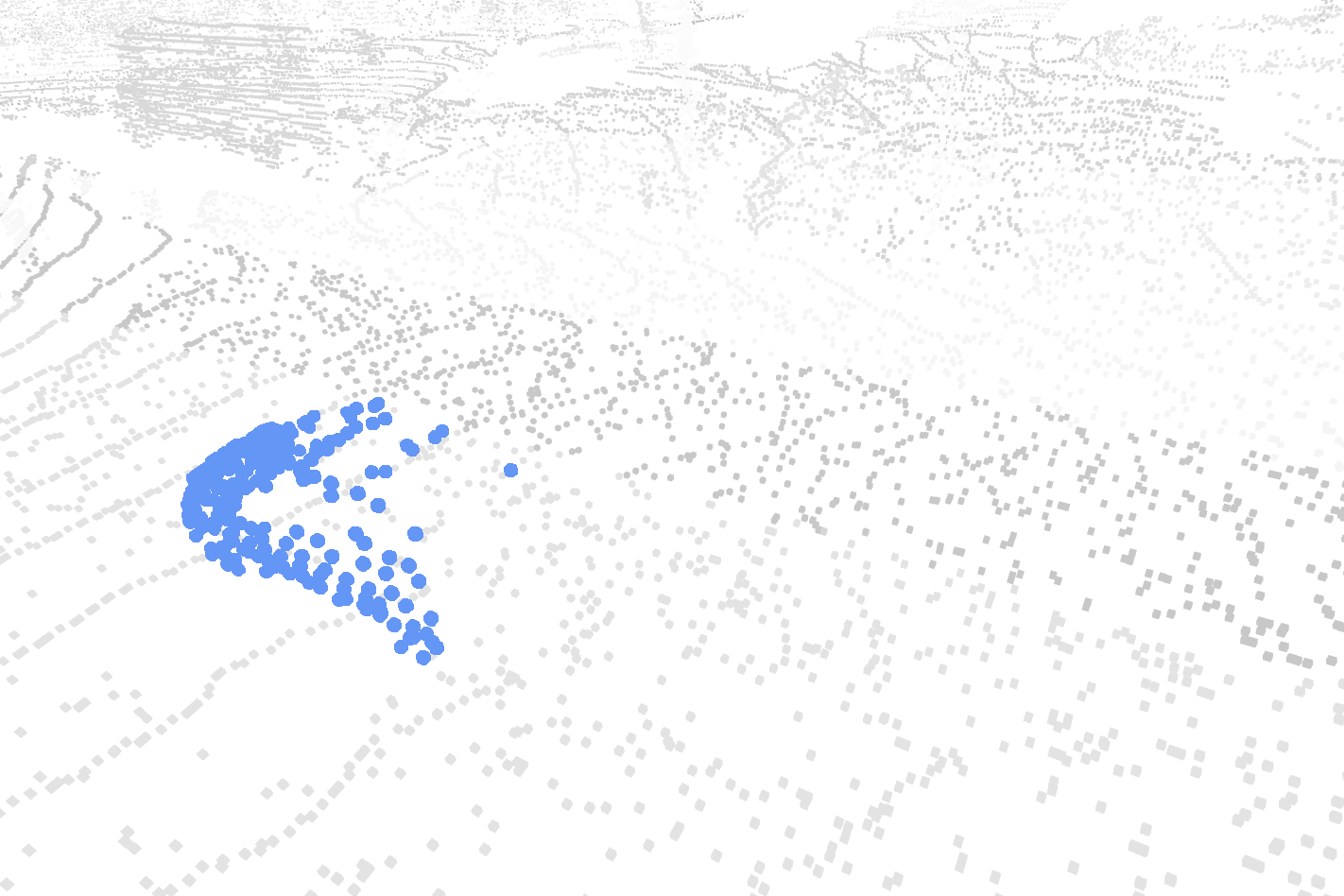}}
    \caption{non-smearing}
  \end{subfigure}%
  \\
  \begin{subfigure}{0.40\textwidth}
  \setlength{\fboxrule}{0.5pt}
    \framebox{\includegraphics[width=.9\linewidth]{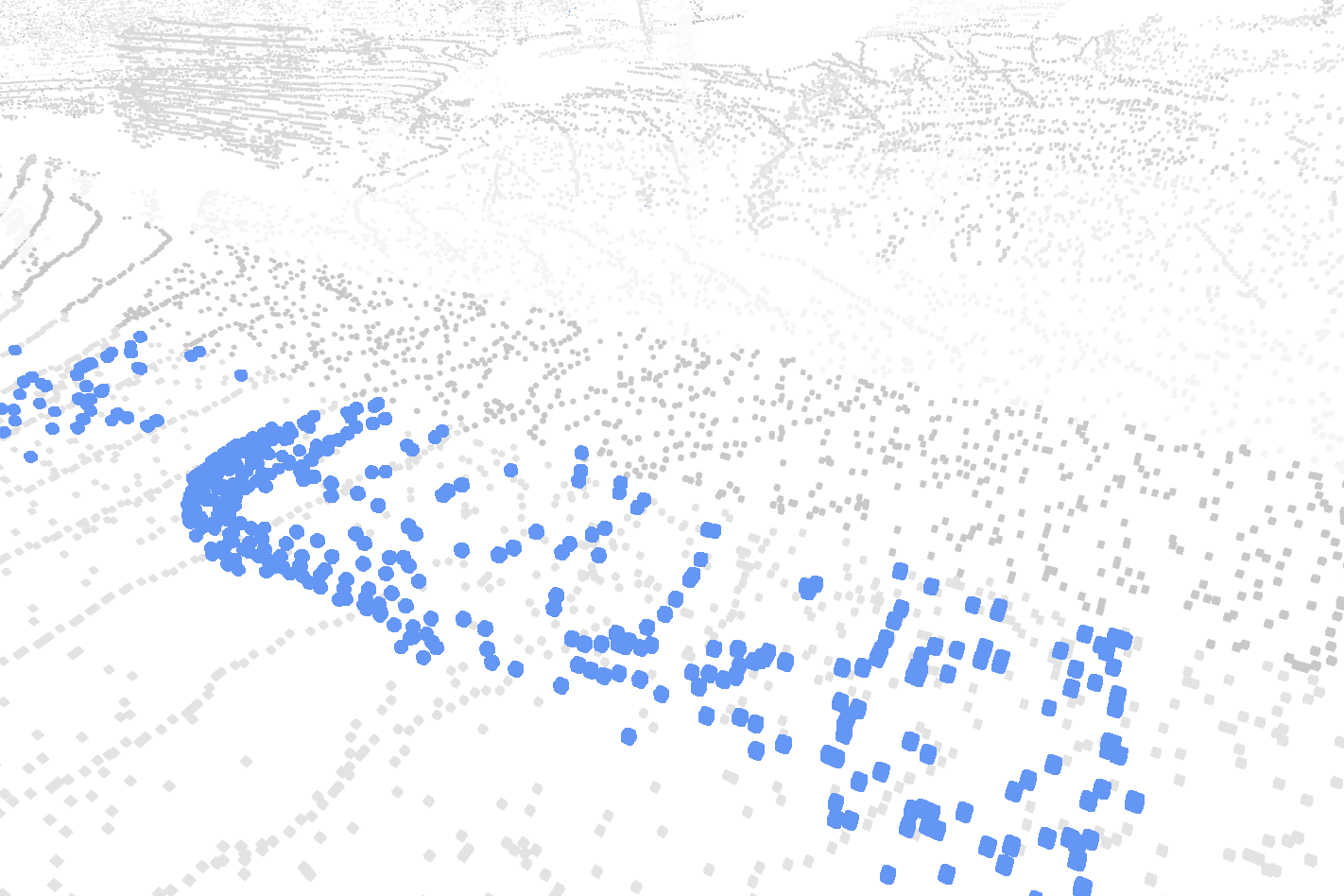}}
    \caption{smearing}
  \end{subfigure}%
\caption{Accumulation (from past and future) of a moving car with the use of ground truth moving object segmentation (a) and with a complete aggregation (b).}
\label{fig:smearing}
\end{figure}

\paragraph{The accumulation of point clouds via egomotion and moving object segmentation}
Once the windows of point clouds are computed, for each window, the egomotion is applied to the corresponding point clouds, in order to accumulate the points with respect to the reference point cloud. Indeed, a pose is associated to each point cloud, in the form of a rotation-translation $4\times 4$ matrix $E$, representing the egomotion with respect to the starting position of the ego-vehicle i.e., 
\begin{equation}
E = 
\begin{bmatrix}
r_{11} & r_{12} & r_{13} & t_{x}\\
r_{21} & r_{22} & r_{23} & t_{y}\\
r_{31} & r_{32} & r_{33} & t_{z}\\
0      & 0      & 0      & 1

\end{bmatrix} \text{,}
\end{equation}
\noindent
where $r_{ij}$ represents the rotations and $t_{x}, t_{y}, t_{z}$ the translations.\\
Given a window of point clouds and the corresponding reference point cloud, we define $E^{*}$ as the matrix associated to the egomotion of the reference point cloud.
Moreover, given a point cloud in the window, we define its egomotion as $E'$.
For each point $p$ of the given point cloud, written as a $4\times 1$ vector,
\begin{equation}
p = 
\begin{bmatrix}
x_{p} \\
y_{p} \\
z_{p} \\
1

\end{bmatrix} ,
\end{equation}
\noindent
the operations in order to perform the accumulation of $p$, with respect to the reference point cloud, are as follows:

\begin{equation}
p'= E' \cdot p \;\text{,}
\end{equation} 

\begin{equation}
p^{*}= E^{*-1} \cdot p' \;\text{,}
\end{equation}
where $p'$ denotes a point of the given point cloud, projected with respect to the starting position of the ego-vehicle, and $p^{*}$ denotes a point of the given point cloud, accumulated in the reference point cloud.
The poses, which are needed to apply the egomotion, can be obtained physically from the host data of the ego-vehicle or numerically via point cloud registration algorithms \cite{Vizzo_2023}, for instance. 
It is important to highlight that moving objects in the environment can not be correctly egomotion-compensated because their movement is not consistent with respect to the egomotion of the ego-vehicle. Therefore, the aggregation of moving objects would cause smearing in the new multi-scan point cloud, like in case of a moving car in Fig. \ref{fig:smearing}. For these reasons, before applying the egomotion-compensation, a moving object segmentation model can be used in the whole sequence in order to identify moving objects. 
The use of a moving object segmentation network is reasonable because the whole preprocessing technique is based on the whole sequence.
Furthermore, several moving object segmentation models were developed in recent years \cite{huang2022dynamic, Mersch_2023}.
Once the detection of moving objects is available, all the points belonging to them can be removed in any point

\noindent
cloud of the window, before the accumulation with respect to the reference point cloud, like in Fig. \ref{fig:smearing}. Both the cases of moving object segmentation (non-smearing) and aggregation of moving objects (smearing) will be analyze in the next section.

\paragraph{A point-based downsampling method}
The architectures considered in this work for the semantic segmentation task are voxel-based models. In those architectures, the 3D space occupied by the point cloud is divided into a regular grid of small 3D cells called voxels and all the points falling in the same voxel are aggregated in order to obtain voxel-based features \cite{choy20194d, graham20173d, graham2017submanifold, zhou2017voxelnet}.
Additionally, all the points residing in the same voxel will be classified the same. 
For this reason, our point-based downsampling method splits the 3D space into the same size as the model, i.e., every cell has the size \texttt{voxel\_size}. Furthermore, a voxel is a reference voxel if it includes at least one point from the reference point cloud (called reference point), and non-reference voxel otherwise.

\begin{algorithm}[!ht]
\DontPrintSemicolon
  
  \KwData{\\ $P \gets$ multi-scan point cloud \\ $P^{*} \gets$ reference point cloud}
  \KwInput{$P$, \texttt{voxel\_size}, \texttt{range}}
  \KwOutput{Downsampled multi-scan point cloud $P'$}
  $V' \gets \{\}$\\
  $V\gets$ Voxelize($P$, \texttt{voxel\_size})\\     
  \For{$v \in V$}{
  
        \If{$ \exists p \in v; \; \exists p \in P^{*} $}{
             $v \gets v \setminus \{p \in v; \; p \notin P^{*}\}$\\
             $V' \gets V' \cup \{v\}$
    
        }
        \Else{
                $p \gets$ ChooseRandom($ p\in v$)\\
                \If{$p$ in \emph{\texttt{range}}}{ 
                    $v \gets \{p\}$\\
                    $V' \gets V' \cup \{v\}$
                }
        }
    }
   $P' \gets \{p' \in v'; \; v' \in V'\}$\\
  \Return{$P'$}\;
\caption{Point-based downsampling}
\label{alg:point}
\end{algorithm}

\noindent
The accumulation, with respect to the reference point cloud, is performed only in the medium and far ranges, by setting a lower and upper range meters. The resulting multi-scan point cloud will be made of reference and non-reference voxels. In order to avoid noise which can be introduced by the accumulation, for every reference voxel we keep only reference points, while for every non-reference voxel, only one random point is kept. The point-based downsampling method is also given in Algo. \ref{alg:point}.

\paragraph{A density-based downsampling method}
In general, voxel-based algorithms are efficient because they rely on the sparsity of voxels. However, when many point clouds are accumulated the sparsity reduces, with a considerable increment in the number of voxels. In addition, aggregating point clouds, each of them with the varying-density property, results in making over dense the closer regions. 
Since the model can deal with only a maximum amount of voxels, an iterative downsampling of voxel is needed. In particular, until the number of voxels with \texttt{voxel\_size} is more than a threshold \texttt{max\_voxel}, the method aggregates voxels for increasing windows of them and donwsamples each window, without removing any reference point. The density-based downsampling method is also given in Algo. \ref{alg:density}.
In this way, thanks to a progressive increment of windows of sparsity, we can achieve a consistent density of voxels along the whole range of accumulation, while limiting the overhead in the total number of voxels.
Additionally, the algorithm removes non-reference points that are more than \texttt{ref\_dist} meters away from any reference point. Indeed, those points do not provide any context for the reference points, as they represent objects that are outside the reference point cloud.
The resulting multi-scan point clouds will have a label for each point in the training set, while for validation and test the labels used are only the ones corresponding to the reference points, in order to compare the performance to single-scan approach.

\begin{algorithm}[!ht]
\DontPrintSemicolon
  
  \KwData{\\ $P' \gets$ multi-scan point cloud returned by Point-based downsampling \\ $P^{*} \gets$ reference point cloud}
  \KwInput{$P'$, \texttt{voxel\_size}, \texttt{max\_voxel}, \texttt{ref\_dist}}
  \KwOutput{Downsampled multi-scan point cloud $P''$}
   
   $V' \gets \{\}$\\
   $V\gets$ Voxelize($P'$, \texttt{ref\_dist})\\
   \For{$v \in V$}{
  
        \If{$ \exists p \in v; \; \exists p \in P^{*} $}{
             $V' \gets V' \cup \{v\}$
        }
    }
    $P' \gets \{p' \in v';\; v' \in V'\}$\\
    $P'' \gets P'$\\

  \texttt{window\_size}$\gets$\texttt{voxel\_size}\\
  \While{True}{

        $V\gets$ Voxelize($P''$, \texttt{voxel\_size})\\     
         \If{$|V| \leq$ \emph{\texttt{max\_voxel}}}{

            \Return{$P''$}
         }
         \Else{

            \texttt{window\_size}$\gets$IncreaseWindow(\texttt{window\_size})\\
            $V'\gets$ Voxelize($P''$, \texttt{window\_size})\\
            $V''\gets$ DownVoxels($V'$, $P^{*}$, \texttt{window\_size})\\
            $P'' \gets \{p'' \in v'';\; v'' \in V''\}$

         }
  }
\caption{Density-based downsampling}
\label{alg:density}
\end{algorithm}

\subsection{SphereFormer} \label{subsec:SphereFormer}

\begin{figure}[ht]
\centering
  \begin{subfigure}{0.5\textwidth}
  \setlength{\fboxrule}{0.5pt}
    \framebox{\includegraphics[width=.9\linewidth]{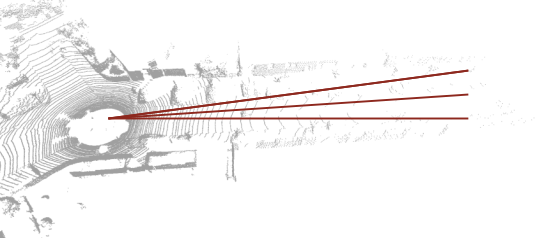}}
  \end{subfigure}%
  \caption{Illustration of the radial window partition.}
   \label{fig:radial_window}
\end{figure}

SphereFormer \cite{lai2023spherical} is one of the first voxel-based models which explicitly addresses the varying-density property by aggregating long-range information directly in a single operator. This is achieved through a module which comprises a radial window self-attention that partitions the space into multiple non-overlapping narrow and long windows. This can overcome the disconnection issue of the sparse distant points and permits to enlarge the receptive field smoothly and dramatically, which significantly boosts the performance of those distant points. Specifically, it represents the 3D space using spherical coordinates $(r, \theta, \phi)$ with the sensor being the origin, and it partitions the scene into multiple non-overlapping windows. Unlike the cubic window shape of vision transformers \cite{vaswani2023attention, fan2021embracing, lai2022stratified, mao2021voxel, sun2022swformer}, the radial windows are long and narrow and they are obtained by partitioning only along the $\theta$ and $\phi$ axis, see Fig. \ref{fig:radial_window}.
The choice of the hyperparameters $\theta$ and $\phi$ is therefore fundamental, as it models the volume of the radial window partitions and hence, the amount of points considered in each radial transformer.
SphereFormer is at the time of writing SOTA model in common benchmarks. Since it is one of the first voxel-based models for outdoor semantic segmentation with the use of a specific vision transformer architecture, for this work it turns out to be a suitable choice. Indeed, transformers permit a bigger receptive field and the self-attention can in principle give some degree of robustness to possible noise introduced with the accumulation of point clouds via egomotion.
The use of a radial transformer architecture is a key ingredient making SphereFormer to a SOTA architecture.
This component was indeed carefully thought for encoding the particular varying-

\noindent 
density property of outdoor semantic segmentation, since it allows to directly aggregate sparse distant points to the denser close ones.
However, the new dataset generated, after the preprocessing step, partially solves the varying-density property by adding voxels in the medium and far ranges.
This overhead in the number of voxels, showed a limitation in the radial transformer. 
Indeed, it turns out that the model with the optimal $\theta$ and $\phi$ for single-scan, is not able to manage an higher number of voxels, due to memory explosion and computational errors. 
However, by decreasing those hyperparameters and hence, decreasing the volume of the radial windows, the model is able to deal with the increased number of voxels.

\subsection{Postprocessing}\label{subsec:Postprocessing}
As explained above, the multi-scan model incorporates a sub-optimal choice of the hyperparameters $\theta$ and $\phi$. Moreover, the whole preprocessing method aims to improve the performances in medium and far ranges only. For these reasons, the performance of the single-scan model in close range is expected to be an upper bound in the performance of the multi-scan model in close range. In order to have an overall improvement of $\mIoU$, the first step of the postprocessing method combines both models in an ensemble modeling fashion, by merging the predictions in close range of the single-scan model with the predictions in medium and far ranges of the multi-scan model.
Furthermore, during validation, we compare the model output with the ground truth for reference points only. However, the multi-scan model can in principle assign a label for all points in any multi-scan point cloud. Therefore, for each multi-scan point cloud of a given sequence, also non-reference points will get a label, and those points will be reference points in another timestamp of the same sequence. Consequently, in the validation of a whole sequence, the same point may be classified multiple times, with different viewpoints. For these reasons, the last step of the postprocessing method performs a sequence-wise classification. In particular, for each point of the sequence, it aggregates all its classifications and via a range-based weight, it assigns to the point the most frequent class.\\
For a point $p$,
\begin{equation}
p = 
\begin{bmatrix}
x \\
y \\
z 

\end{bmatrix}\text{,}
\end{equation}

\noindent
we can define its radius r as
\begin{equation}
r = \sqrt{x^{2} + y^{2} + z^{2}}\;\text{,}
\end{equation}

\noindent
and we can define its weight $w_{r}$ as
\begin{equation}
 w_{r} = \max \left( 0.1, \dfrac{20\text{m}}{r + 20\text{m}}\right)\;\text{.}
\end{equation}
In this way, the closer the point is, the higher the weight. In particular, this function maps points with a radius $r \approx 0 $m near to $1$, since those points are very close to the ego-vehicle and therefore are very reliable, and clips points with a radius  $r \geq 180 $m to $0.1$ because we want to rely lightly in their classification. Generally speaking, this function scales smoothly for increasing radius and provides reasonable weights $w_r$ reflecting the point densities for different ranges. Indeed, breaking down the radius in the three considered ranges, we get approximately intervals of $w_r \in [1, 0.5[$ for close range ($r <20$m), $w_r \in [0.5, 0.3[$ for medium range ($20$m $\leq r$ $<50$m) and $w_r \in [0.3, 0.1]$ for far range ($r \geq 50$m).

\section{Experiments} \label{sec:results}

For experiments we used two datasets: SemanticKITTI \cite{behley2019semantickitti} and nuScenes \cite{caesar2020nuscenes}. In particular, referring to Sec. \ref{subsec:preprocessing}, for each dataset we generated two multi-scan datasets. In one we removed moving objects using the ground truth available in the datasets: prep (non-smearing), and in the other we performed the preprocessing without removing any moving object: prep (smearing), in order to analyze differences in the performance.\\
First, we describe the settings of the preprocessing for both datasets and afterwards, we conduct an analysis of the performance on semantic segmentation.
All the results for each model are obtained as the top performance among three runs of the same experiment and they consist of a single forward pass on the validation set\footnote{The computations were carried out on the PLEIADES cluster at the University of Wuppertal, which was supported by the Deutsche Forschungsgemeinschaft (DFG, grant No. INST 218/78-1 FUGG) and the Bundesministerium für Bildung und Forschung (BMBF).}.
We decided to pick the top performance models because the variability on the results depends mostly on the performance in close range. This does not affect our experiments because our method focuses in medium and far ranges and the postprocessing stage merges the close range performances. We also decided to not report the variability on the results as this strictly depends on the downstream voxel-based model and not on our method. Indeed, the preprocessing procedure affects only the dataset and the postprocessing is a deterministic procedure which deals with already given models.

\subsection{SemanticKITTI} \label{subsec:kitti}

The SemanticKITTI dataset \cite{behley2019semantickitti} contains street scenes from and around Karlsruhe, Germany. It provides $11$ sequences with about $23$K samples for training and validation, consisting of $19$ classes. The LiDAR sensor has $64$ channels. Furthermore the data is recorded and annotated with $10$Hz and each point cloud contains about $120$K points.
The authors of the dataset recommend to use all sequences to train the LiDAR segmentation model, except sequence $08$, which should be used for validation.
Following the settings of our downstream voxel-based semantic segmentation model \cite{lai2023spherical}, we set $\texttt{voxel\_size}=0.05$m.
 The choice of \texttt{accumulate\_length} is $20$, for a good trade off between the time needed for the preprocessing and the quality of density in the medium and far regions. With an accumulation of $20$ point clouds,  \texttt{min\_dist} is set to $2$m, with a meaning of having in total around $\texttt{accumulate\_length} * \texttt{min\_dist} = 40$m of translation (in past and future) with respect to the reference point cloud, on which the scanned region is dense. For example, if in the reference point cloud a not moving object was distant $40$m with respect to the ego-vehicle  (medium range), there will exist at least one selected point cloud in which the ego-vehicle will be close to such object. Therefore, it will be possible to aggregate a sufficient amount of points coming from the closer and denser region, belonging to such an object, in the reference point cloud, where the object was initially underrepresented. This range of translation is reasonable for SemanticKITTI as its segmentation is limited to a range of $\sim50$m.
 Furthermore, given the number of voxels in a single-scan $\sim120$K, a straightforward accumulation of $\texttt{accumulate\_length}=20$ point clouds would lead to over $2$M voxels. However, by applying our preprocessing, described in Sec. \ref{subsec:preprocessing}, we set the maximum number of voxels for the multi-scan generation to $\texttt{max\_voxel}=180$K for a good trade off between the aggregation quality and the model capabilities. Additionally, \texttt{ref\_dist} is set to $5$m since all the objects of interest present dimension inferior to $5$m.
 Overall, the voxel distribution of SemanticKITTI is on average $74\%$ in close, $22\%$ in medium and $4\%$ in far ranges. After the preprocessing, the voxel distribution becomes on average $50\%$ in close, $40\%$ in medium and $10\%$ in far ranges.
 
\subsection{nuScenes}

The nuScenes dataset \cite{caesar2020nuscenes} contains street scenes from two cities, Boston (US) and Singapore. It provides $700$ sequences for training and $150$ sequences for validation, consisting of $16$ classes. Each sequence contains about $40$ samples which amounts to a total of $34$K key frames. The LiDAR sensor has $32$ channels. Furthermore, the dataset is recorded and annotated with $2$Hz and each point cloud contains roughly $40$K points.
Following the settings of our downstream voxel-based semantic segmentation model \cite{lai2023spherical}, we set $\texttt{voxel\_size}=0.1$m.
The choice of \texttt{accumulate\_length} is $40$ (which corresponds to each whole sequence), because of the sparser resolution given by the $32$ channels. Still \texttt{min\_dist} is set to $2$m, with a meaning of having in total around $\texttt{accumulate\_length} * \texttt{min\_dist} = 80$m of translation (in past and future) with respect to the reference point cloud, on which the scanned region is dense. This range of translation is reasonable for nuScenes as its segmentation is limited to a range of $\sim120$m.
Furthermore, given the number of voxels in a single-scan $\sim40$K, a straightforward accumulation of $\texttt{accumulate\_length}=40$ point clouds would lead to over $1$M voxels. However, by applying our preprocessing, described in Sec. \ref{subsec:preprocessing}, we set the maximum number of voxels for the multi-scan generation to $\texttt{max\_voxel}=120$K for a good trade off between the aggregation quality and the model capabilities. In particular, the number of voxels was chosen lower than on SemanticKITTI because of the sparser resolution of nuScenes combined with its larger semantic segmentation range. Additionally, \texttt{ref\_dist} is set to $5$m since all the objects of interest present dimension inferior to $5$m.
Overall, the voxel distribution of nuScenes is on average $72\%$ in close, $22\%$ in medium and $6\%$ in far ranges. After the preprocessing, the voxel distribution becomes on average $40\%$ in close, $40\%$ in medium and $20\%$ in far ranges.

\subsection{Semantic Segmentation Results}

\begin{figure}[h]
\centering
\begin{subfigure}{0.51\textwidth}
  \setlength{\fboxrule}{0.5pt}
    \framebox{\includegraphics[width=.9\linewidth]{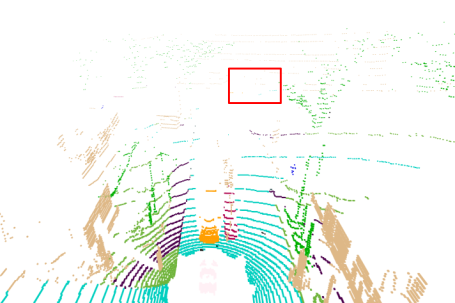}}
    \caption{Overall scene}
  \end{subfigure}%
  \\
  \begin{subfigure}{0.24\textwidth}
  \setlength{\fboxrule}{0.5pt}
    \framebox{\includegraphics[width=.9\linewidth]{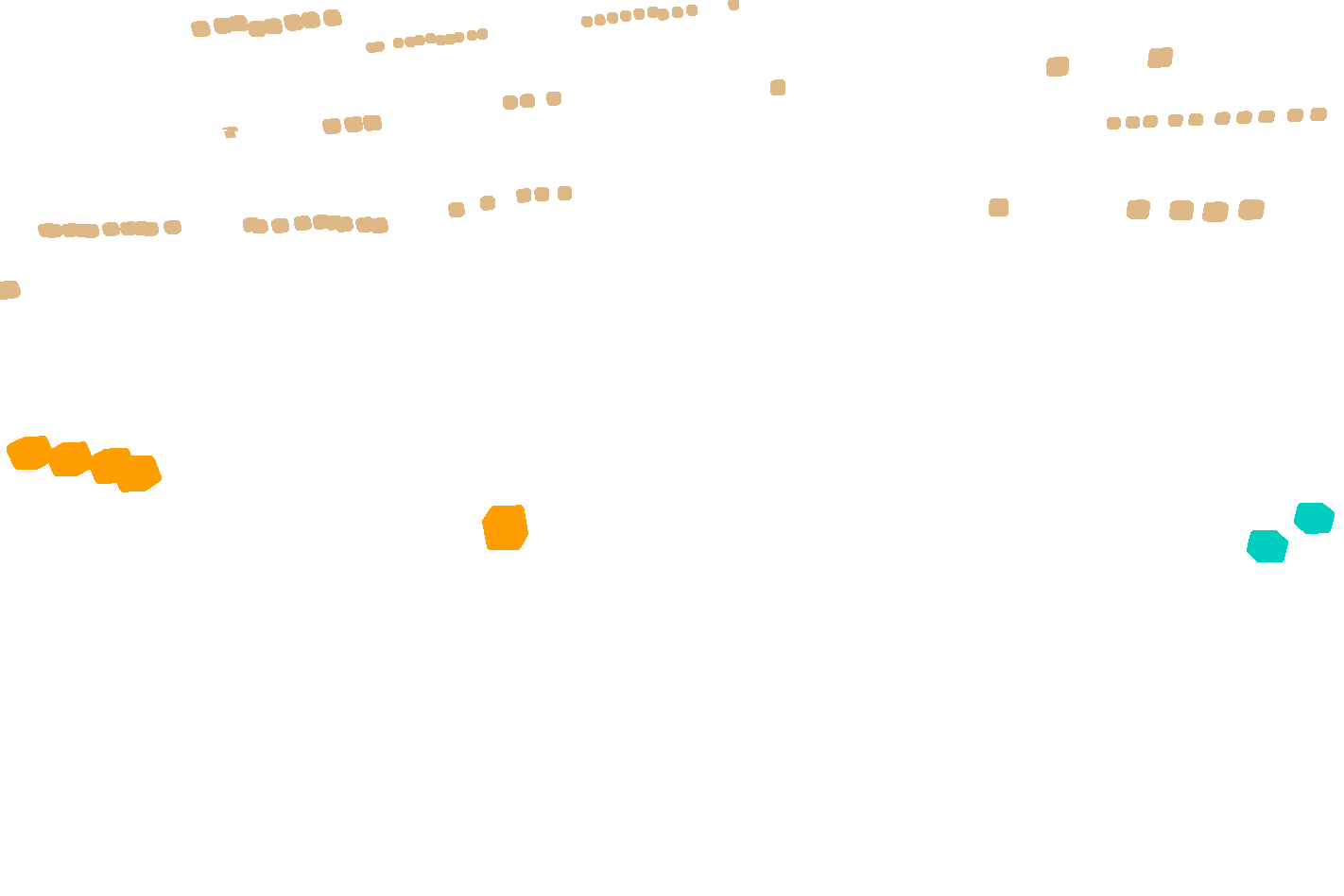}}
    \caption{Ground truth}
  \end{subfigure}%
  \begin{subfigure}{0.24\textwidth}
  \setlength{\fboxrule}{0.5pt}
    \framebox{\includegraphics[width=.9\linewidth]{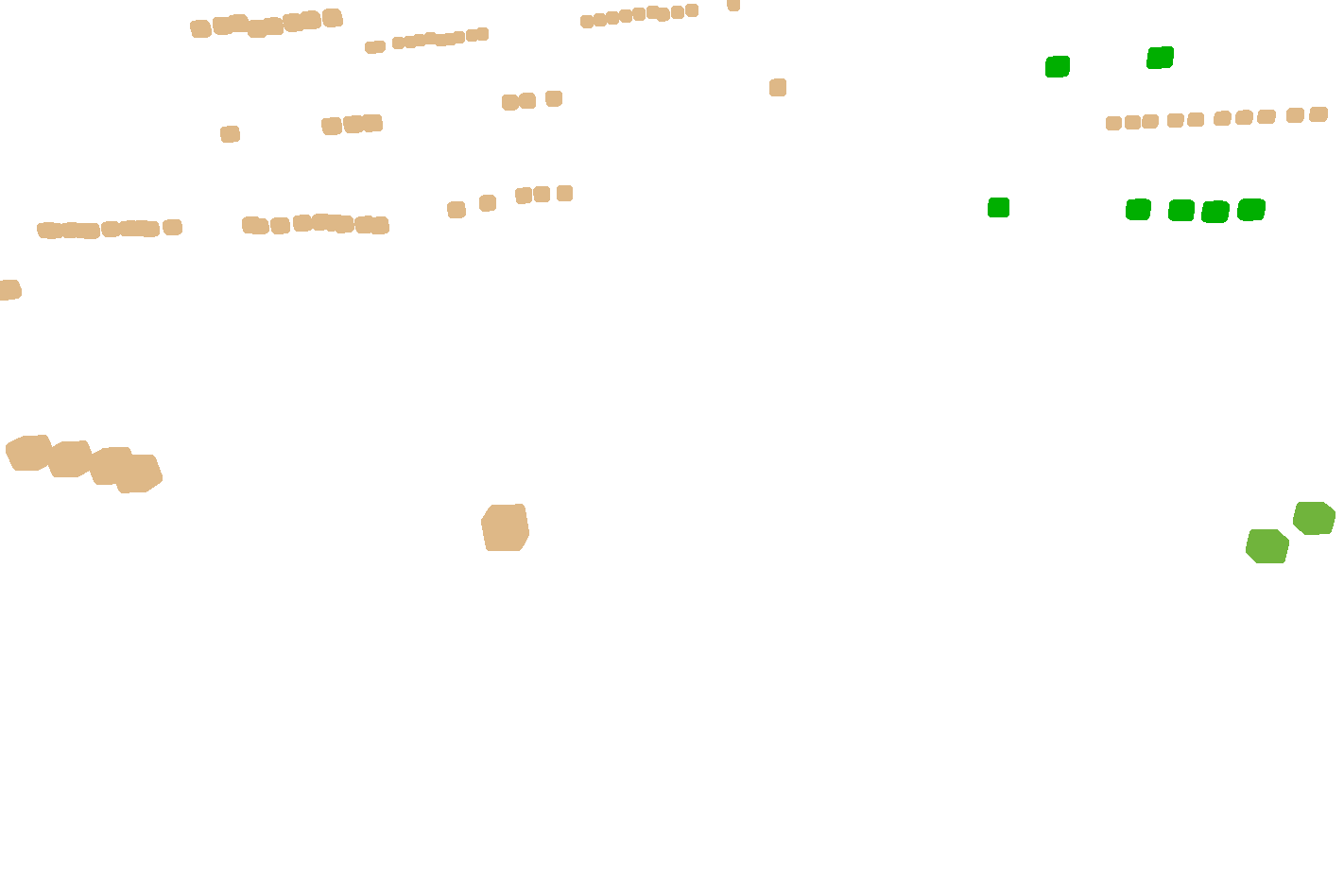}}
    \caption{Prediction of single-scan}
  \end{subfigure}%
  \\
  \begin{subfigure}{0.24\textwidth}
  \setlength{\fboxrule}{0.5pt}
    \framebox{\includegraphics[width=.9\linewidth]{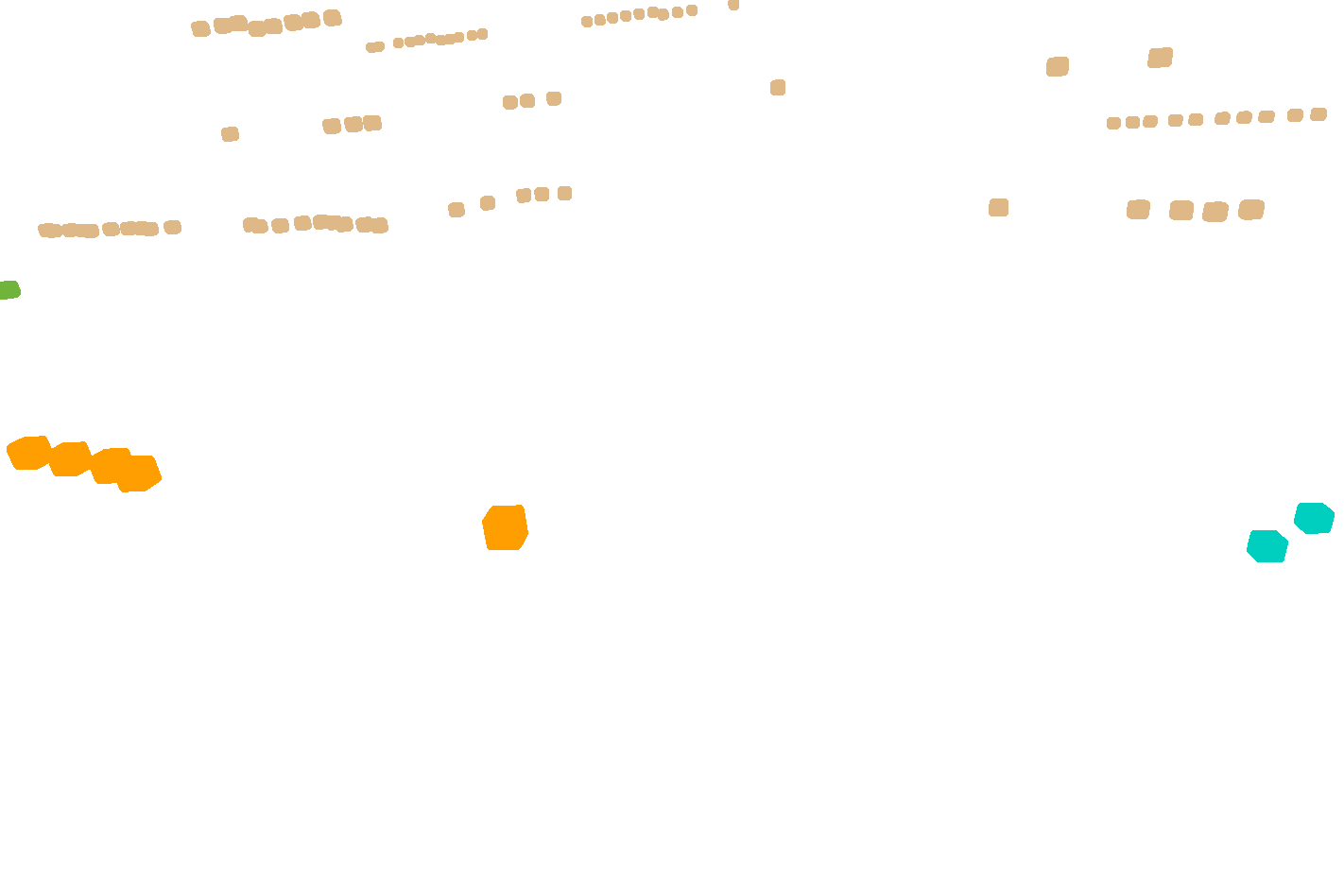}}
    \caption{Prediction of our method}
  \end{subfigure}%
  \begin{subfigure}{0.24\textwidth}
  \setlength{\fboxrule}{0.5pt}
    \framebox{\includegraphics[width=.9\linewidth]{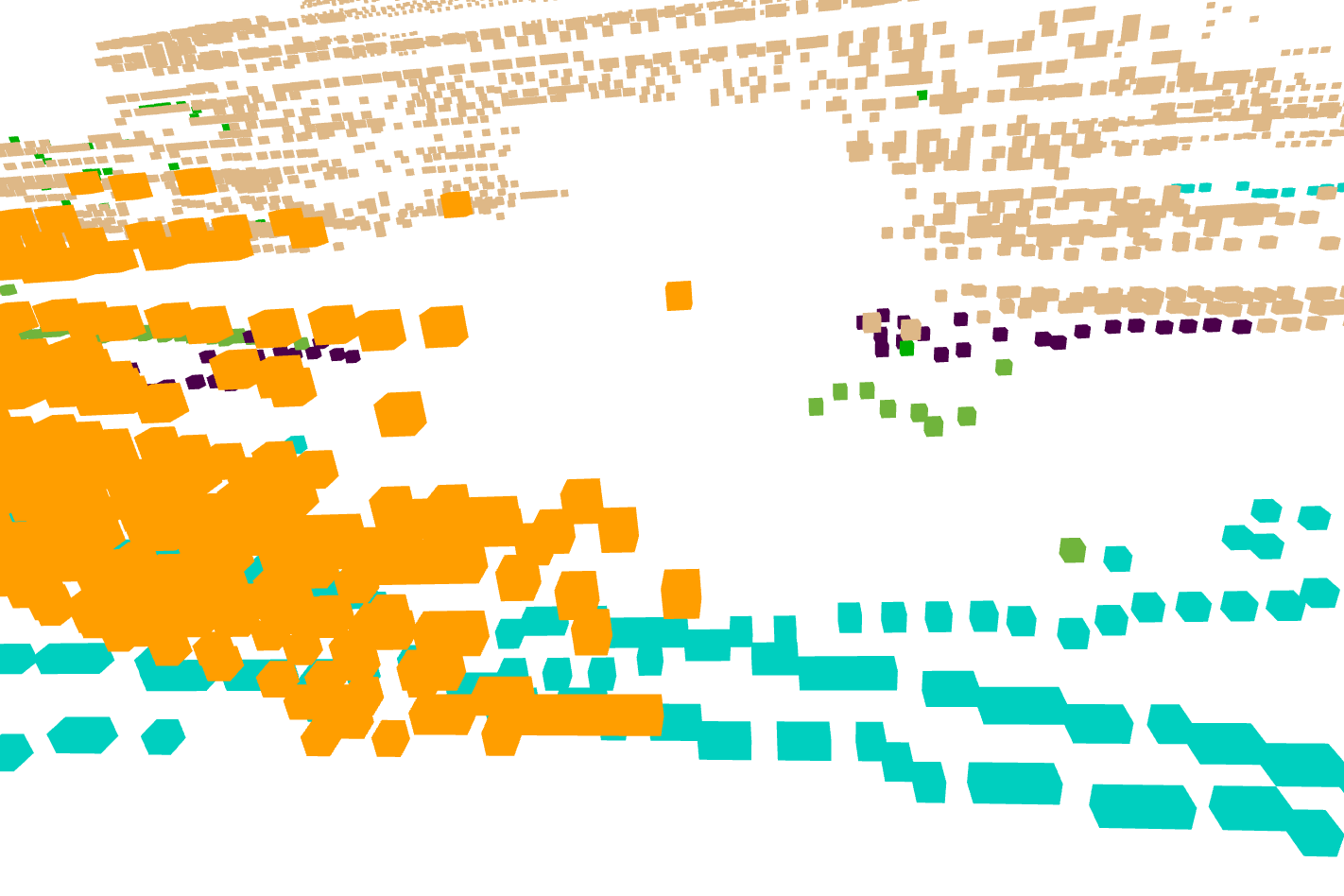}}
    \caption{Aggregation of our method}
  \end{subfigure}%
\caption{Semantic segmentation performance in far range on nuScenes: (b)-(e) represent the area inside the red rectangle in (a). Due to the sparsity in long distance, the single-scan prediction (c) misclassified "vehicle.car" - orange points in (b) - ; "flat.driveable\_surface" - light blue in (b) - ; "static.other" - bisque points in  (b) - respectively with "static.other" - bisque points in  (c) - ; "flat.terrain" - light green in (c) - and "static.vegetation" - green in (c) - .}
\label{fig:preds}
\end{figure}

The results on SemanticKITTI val set are shown in Tables \ref{semkitti_miou}, \ref{semkitti_classes}. While single-scan $\text{model}_{s}$ achieves $67.5\%$ $\mIoU$, our method $\text{prep (non-smearing)} + \text{model}_{m} + \text{postp}$ achieves $68.3\%$ $\mIoU$. Both the preprocessing methods, $\text{prep (non-smearing)} + \text{model}_{m}$ and $\text{prep (smearing)} \,+\, \text{model}_{m}$, outperform the single-scan $\text{model}_{s}$ in medium range of over $4$ percentage point. However, in far range the preprocessing methods do not consistently improve the performance. In fact SphereFormer clips the point clouds during the processing of SemanticKITTI to $51.2$m, as the segmentation of SemanticKITTI is limited to a range of $\sim50$m. Consequently, the accumulation in far range, which is defined for $\geq 50$m, in this specific case only accounts $1.2$m. This means that for SemanticKITTI the accumulation in far range is not effective because most of the objects can not benefit from the accumulation, as the clipping would not permit to give a defined structure to those far objects. Furthermore, in close range there is a degradation of the performance. This is well explained by the fact that the multi-scan model turns out to be sub-optimal (because of the use of sub-optimal hyperparameters $\theta$ and $\phi$), as remarked in Subsec. \ref{subsec:Postprocessing}, combined with the preprocessing method which only aims to improve results in long distance. The postprocessing method, $\text{prep (non-smearing)} \,+\, \text{model}_{m} \,+\, \text{postp}$ and $\text{prep (smearing)} \,+\, \text{model}_{m} \,+\, \text{postp}$, permits to bring the multi-scan performance in close range again comparable to single-scan, thanks to the ensemble modeling. Additionally, it improves $\mIoU$ in the overall range of $0.8$ percentage point thanks to the voting scheme. In particular, it increases the performance in medium range from $58.8\%$ $\mIoU$ of $\text{model}_{s}$ to $64.6\%$ $\mIoU$ of $\text{prep (non-smearing)} \,+\, \text{model}_{m} \,+\, \text{postp}$ (over $5$ percentage point improvement), and it drastically improves in far range from $16.2\%$ $\mIoU$ of $\text{model}_{s}$ to $25.4\%$ $\mIoU$ of $\text{prep (smearing)} \,+\, \text{model}_{m} \,+\, \text{postp}$ (over $9$ percentage point improvement). The improvement is particularly evident for example in class "parking", where from $50.8\%$ $\mIoU$ in medium and $20.6\%$ $\mIoU$ in far ranges of $\text{model}_{s}$ it reaches respectively $64.4\%$ $\mIoU$ and $45.9\%$ $\mIoU$ in $\text{prep (smearing)} \,+\, \text{model}_{m} \,+\, \text{postp}$.
These significant increments of performance are however not well reflected in the overall  $\mIoU$, due to the varying-density property of the points. In Table \ref{semkitti_classes_test} are also shown the results on SemanticKITTI test set. It is worth to mention that only prep (smearing) was utilized in the test set, since the non-smearing procedure was based on ground truth, not available for the testing procedure. The overall $\mIoU$ improvement of $0.7$ percentage point is consistent with the performance of the same method on SemanticKITTI val set and proves the generalization capabilities of our method.

\begin{table}[hbt!]
\begin{center}
\caption{Range-based semantic segmentation results on SemanticKITTI val set. The points are on average distributed as $74\%$ in close, $22\%$ in medium and $4\%$ in far ranges.}
\label{semkitti_miou}
\scalebox{0.9}{
\begin{tabular}{l|l|lll}
\hline
Method & mIoU & close & medium & far
\\ \hline
$\text{model}_{s}$ & $67.5$ & $\mathbf{68.5}$ & $58.8$ & $16.2$ \\
$\text{prep (non-smearing)} + \text{model}_{m}$ & $68.1$ & $68.3$ & $63.7$ & $15.5$ \\
$\text{prep (smearing)} + \text{model}_{m}$ & $67.4$ & $67.8$ & $62.4$ & $17.6$ \\
$\text{prep (non-smearing)} + \text{model}_{m} + \text{postp}$ & $\mathbf{68.3}$ & $\mathbf{68.5}$ & $\mathbf{64.6}$ & $23.4$ \\
$\text{prep (smearing)} + \text{model}_{m} + \text{postp}$ & $68.2$ & $\mathbf{68.5}$ & $63.3$ & $\mathbf{25.4}$ \\
\end{tabular}
}
\end{center}
\end{table}

\begin{table*}[hbt!]
\begin{center}
\captionsetup{justification=centering}
\caption{Semantic segmentation results on SemanticKITTI val set overall and split on close, medium, far ranges.}
\label{semkitti_classes}
\scalebox{0.65}{
\begin{tabular}{|l|l|l|llllllllllllllllllr|}
\cline{3-22}
\multicolumn{2}{c|}{} & mIoU  & \rotatebox[origin=c]{90}{car} & \rotatebox[origin=c]{90}{bicycle} & \rotatebox[origin=c]{90}{motorcycle} & \rotatebox[origin=c]{90}{truck} & \rotatebox[origin=c]{90}{other-veh.} & \rotatebox[origin=c]{90}{person} & \rotatebox[origin=c]{90}{bicyclist} & \rotatebox[origin=c]{90}{ motorcyclist } & \rotatebox[origin=c]{90}{road} & \rotatebox[origin=c]{90}{parking} & \rotatebox[origin=c]{90}{sidewalk} & \rotatebox[origin=c]{90}{other-gro.} & \rotatebox[origin=c]{90}{building} & \rotatebox[origin=c]{90}{fence} & \rotatebox[origin=c]{90}{vegetation} & \rotatebox[origin=c]{90}{trunk} & \rotatebox[origin=c]{90}{terrain} & \rotatebox[origin=c]{90}{pole} & \rotatebox[origin=c]{90}{traffic sign}
\\
\hline
\multirow{5}{*}{\rotatebox{90}{overall}}
 & $\text{model}_{s}$ & $67.5$  & $96.9$ & $51.8$ & $73.7$ & $84.7$ & $69.0$ & $\mathbf{76.3}$ & $92.2$ & $0.0$ & $94.9$ & $51.9$ & $82.5$ & $10.5$ & $90.0$ & $57.0$ & $88.7$ & $71.1$ & $75.6$ & $63.8$ & $52.2$ \\
 & $\text{prep (non-smearing)} + \text{model}_{m}$ & $68.1$  & $96.6$ & $49.4$ & $\mathbf{79.1}$ & $81.9$ & $61.5$ & $74.0$ & $91.0$ & $0.0$ & $93.9$ & $\mathbf{58.2}$ & $81.4$ & $7.9$ & $\mathbf{92.0}$ & $\mathbf{67.2}$ & $\mathbf{89.9}$ & $\mathbf{72.9}$ & $\mathbf{78.1}$ & $65.9$ & $53.0$ \\
 & $\text{prep (smearing)} + \text{model}_{m}$ & $67.4$  & $96.5$ & $45.7$ & $76.4$ & $\mathbf{87.1}$ & $59.1$ & $76.1$ & $91.6$ & $0.0$ & $94.8$ & $54.8$ & $82.3$ & $9.4$ & $90.5$ & $57.6$ & $89.1$ & $70.4$ & $76.7$ & $\mathbf{67.4}$ & $\mathbf{54.3}$ \\
 & $\text{prep (non-smearing)} + \text{model}_{m} + \text{postp}$ & $\mathbf{68.3}$  & $\mathbf{97.0}$ & $\mathbf{52.1}$ & $75.5$ & $87.0$ & $\mathbf{70.4}$ & $76.2$ & $92.9$ & $0.0$ & $95.0$ & $52.1$ & $82.6$ & $12.8$ & $90.1$ & $57.8$ & $89.1$ & $72.2$ & $76.3$ & $65.2$ & $54.1$ \\
 & $\text{prep (smearing)} + \text{model}_{m} + \text{postp}$ & $68.2$  & $96.9$ & $52.0$ & $75.3$ & $84.8$ & $69.7$ & $76.0$ & $\mathbf{93.1}$ & $0.0$ & $\mathbf{95.1}$ & $53.0$ & $\mathbf{82.8}$ & $\mathbf{13.1}$ & $90.2$ & $57.1$ & $88.9$ & $72.2$ & $76.0$ & $65.0$ & $53.8$ \\
\hline
\multirow{5}{*}{\rotatebox{90}{close}}
 & $\text{model}_{s}$ & $\mathbf{68.5}$  & $\mathbf{97.2}$ & $54.7$ & $75.9$ & $86.7$ & $\mathbf{70.4}$ & $78.4$ & $94.6$ & $0.0$ & $\mathbf{95.2}$ & $52.0$ & $\mathbf{83.5}$ & $\mathbf{12.4}$ & $89.6$ & $58.7$ & $88.8$ & $73.5$ & $76.2$ & $65.4$ & $48.2$ \\
 & $\text{prep (non-smearing)} + \text{model}_{m}$ & $68.3$  & $96.8$ & $51.4$ & $\mathbf{79.8}$ & $81.2$ & $60.7$ & $76.1$ & $92.8$ & $0.0$ & $94.1$ & $\mathbf{58.9}$ & $82.3$ & $6.3$ & $\mathbf{91.9}$ & $\mathbf{68.9}$ & $\mathbf{89.9}$ & $\mathbf{74.7}$ & $\mathbf{78.4}$ & $66.6$ & $47.4$ \\
 & $\text{prep (smearing)} + \text{model}_{m}$ & $67.8$  & $96.8$ & $47.7$ & $77.3$ & $\mathbf{89.5}$ & $58.9$ & $\mathbf{78.8}$ & $93.4$ & $0.0$ & $95.0$ & $54.1$ & $83.0$ & $7.9$ & $89.9$ & $59.4$ & $89.1$ & $71.5$ & $77.2$ & $\mathbf{68.4}$ & $\mathbf{49.7}$ \\
& $\text{prep (non-smearing)} + \text{model}_{m} + \text{postp}$ & $\mathbf{68.5}$  & $\mathbf{97.2}$ & $\mathbf{54.8}$ & $75.9$ & $86.7$ & $\mathbf{70.4}$ & $78.4$ & $94.6$ & $0.0$ & $\mathbf{95.2}$ & $52.0$ & $\mathbf{83.5}$ & $\mathbf{12.4}$ & $89.6$ & $58.7$ & $88.8$ & $73.5$ & $76.2$ & $65.4$ & $48.3$ \\
 & $\text{prep (smearing)} + \text{model}_{m} + \text{postp}$ & $\mathbf{68.5}$  & $\mathbf{97.2}$ & $54.7$ & $75.9$ & $86.7$ & $\mathbf{70.4}$ & $78.5$ & $\mathbf{94.8}$ & $0.0$ & $\mathbf{95.2}$ & $52.0$ & $\mathbf{83.5}$ & $\mathbf{12.4}$ & $89.6$ & $58.7$ & $88.8$ & $73.5$ & $76.2$ & $65.4$ & $48.3$ \\
\hline
\multirow{5}{*}{\rotatebox{90}{medium}}
 & $\text{model}_{s}$ & $58.8$  & $92.1$ & $32.1$ & $47.7$ & $73.4$ & $57.8$ & $\mathbf{65.2}$ & $69.6$ & $0.0$ & $90.2$ & $50.8$ & $67.4$ & $4.3$ & $91.9$ & $35.8$ & $88.2$ & $63.8$ & $71.4$ & $55.5$ & $60.6$ \\
 & $\text{prep (non-smearing)} + \text{model}_{m}$ & $63.7$  & $93.5$ & $35.6$ & $68.8$ & $86.4$ & $70.3$ & $62.8$ & $73.1$ & $0.0$ & $90.9$ & $50.6$ & $67.1$ & $13.7$ & $92.4$ & $44.1$ & $90.1$ & $67.2$ & $75.9$ & $62.6$ & $65.8$ \\
 & $\text{prep (smearing)} + \text{model}_{m}$ & $62.4$  & $92.4$ & $31.7$ & $65.2$ & $73.2$ & $62.4$ & $62.8$ & $73.7$ & $0.0$ & $92.9$ & $62.3$ & $70.6$ & $14.4$ & $92.9$ & $36.0$ & $89.0$ & $67.0$ & $73.0$ & $61.8$ & $65.2$ \\
 & $\text{prep (non-smearing)} + \text{model}_{m} + \text{postp}$ & $\mathbf{64.6}$  & $\mathbf{93.7}$ & $\mathbf{36.1}$ & $\mathbf{70.2}$ & $\mathbf{88.6}$ & $\mathbf{71.2}$ & $64.4$ & $75.1$ & $0.0$ & $91.1$ & $52.4$ & $67.8$ & $14.0$ & $92.3$ & $\mathbf{45.0}$ & $\mathbf{90.4}$ & $\mathbf{68.0}$ & $\mathbf{76.9}$ & $\mathbf{63.7}$ & $\mathbf{66.7}$ \\
 & $\text{prep (smearing)} + \text{model}_{m} + \text{postp}$ & $63.3$  & $92.4$ & $32.7$ & $66.4$ & $73.3$ & $63.0$ & $63.5$ & $\mathbf{76.5}$ & $0.0$ & $\mathbf{93.2}$ & $\mathbf{64.4}$ & $\mathbf{71.7}$ & $\mathbf{15.1}$ & $\mathbf{93.0}$ & $37.0$ & $89.3$ & $67.7$ & $74.1$ & $62.8$ & $66.0$ \\
\hline
\multirow{5}{*}{\rotatebox{90}{far}}
 & $\text{model}_{s}$ & $16.2$  & $\mathbf{41.7}$ & $0.0$ & $0.0$ & $0.0$ & $8.1$ & $0.0$ & $0.0$ & $0.0$ & $44.4$ & $20.6$ & $10.1$ & $0.0$ & $21.5$ & $2.6$ & $84.0$ & $7.2$ & $27.0$ & $32.1$ & $8.2$ \\
 & $\text{prep (non-smearing)} + \text{model}_{m}$ & $15.5$  & $16.7$ & $0.0$ & $0.0$ & $0.0$ & $9.2$ & $0.0$ & $0.0$ & $0.0$ & $24.0$ & $27.1$ & $13.7$ & $0.0$ & $23.2$ & $6.4$ & $85.9$ & $6.3$ & $32.1$ & $26.8$ & $22.9$ \\
 & $\text{prep (smearing)} + \text{model}_{m}$ & $17.6$  & $31.8$ & $0.0$ & $0.0$ & $0.0$ & $11.9$ & $0.0$ & $0.0$ & $0.0$ & $41.7$ & $32.9$ & $14.8$ & $0.0$ & $24.8$ & $3.3$ & $86.7$ & $6.2$ & $34.0$ & $33.7$ & $13.4$ \\
 & $\text{prep (non-smearing)} + \text{model}_{m} + \text{postp}$ & $23.4$  & $21.7$ & $0.0$ & $0.0$ & $0.0$ & $26.9$ & $0.0$ & $0.0$ & $0.0$ & $37.7$ & $39.5$ & $33.0$ & $0.0$ & $\mathbf{50.4}$ & $\mathbf{11.9}$ & $89.8$ & $\mathbf{8.4}$ & $48.8$ & $41.8$ & $\mathbf{33.8}$ \\
 & $\text{prep (smearing)} + \text{model}_{m} + \text{postp}$ & $\mathbf{25.4}$  & $37.4$ & $0.0$ & $0.0$ & $0.0$ & $\mathbf{27.6}$ & $0.0$ & $0.0$ & $0.0$ & $\mathbf{59.8}$ & $\mathbf{45.9}$ & $\mathbf{34.0}$ & $0.0$ & $49.1$ & $5.9$ & $\mathbf{91.1}$ & $7.6$ & $\mathbf{52.2}$ & $\mathbf{47.5}$ & $24.4$ \\
\hline
\end{tabular}
}
\end{center}
\end{table*}

\begin{table*}[hbt!]
\begin{center}
\caption{Semantic segmentation results on SemanticKITTI test set.}
\label{semkitti_classes_test}
\scalebox{0.65}{
\begin{tabular}{l|l|lllllllllllllllllll}
\hline
Method & mIoU  & \rotatebox[origin=c]{90}{car} & \rotatebox[origin=c]{90}{bicycle} & \rotatebox[origin=c]{90}{motorcycle} & \rotatebox[origin=c]{90}{truck} & \rotatebox[origin=c]{90}{other-veh.} & \rotatebox[origin=c]{90}{person} & \rotatebox[origin=c]{90}{bicyclist} & \rotatebox[origin=c]{90}{ motorcyclist } & \rotatebox[origin=c]{90}{road} & \rotatebox[origin=c]{90}{parking} & \rotatebox[origin=c]{90}{sidewalk} & \rotatebox[origin=c]{90}{other-gro.} & \rotatebox[origin=c]{90}{building} & \rotatebox[origin=c]{90}{fence} & \rotatebox[origin=c]{90}{vegetation} & \rotatebox[origin=c]{90}{trunk} & \rotatebox[origin=c]{90}{terrain} & \rotatebox[origin=c]{90}{pole} & \rotatebox[origin=c]{90}{traffic sign} 
\\ \hline
 $\text{model}_{s}$ & $65.0$ & $95.9$ & $52.0$ & $54.6$ & $49.5$ & $46.2$ & $66.6$ & $63.0$ & $\mathbf{30.5}$ & $91.6$ & $63.5$ & $75.0$ & $\mathbf{35.3}$ & $93.2$ & $70.7$ & $84.5$ & $70.7$ & $67.3$ & $62.3$ & $63.4$ \\
 $\text{prep (smearing)} + \text{model}_{m} + \text{postp}$ & $\mathbf{65.7}$ & $\mathbf{96.1}$ & $\mathbf{52.3}$ & $\mathbf{55.1}$ & $\mathbf{52.0}$ & $\mathbf{47.7}$ & $\mathbf{66.8}$ & $\mathbf{63.6}$ & $30.1$ & $\mathbf{91.7}$ & $\mathbf{63.9}$ & $\mathbf{75.3}$ & $34.7$ & $\mathbf{93.3}$ & $\mathbf{71.2}$ & $\mathbf{84.7}$ & $\mathbf{71.9}$ & $\mathbf{67.6}$ & $\mathbf{63.7}$ & $\mathbf{66.9}$
\end{tabular}
}
\end{center}
\end{table*}

The results on nuScenes val set are shown in Tables \ref{nuscenes_miou}, \ref{nuscenes_classes}. While single-scan $\text{model}_{s}$ achieves $77.8\%$ $\mIoU$, our method $\text{prep (non-smearing)} + \text{model}_{m} + \text{postp}$ achieves $79.4\%$ $\mIoU$.
Also in this case the preprocessing methods, $\text{prep (non-smearing)} \,+\, \text{model}_{m}$ and $\text{prep (smearing)} \,+\, \text{model}_{m}$, outperform the single-scan $\text{model}_{s}$ in medium range of over $5$ percentage point. In contrast to SemanticKITTI, these methods also outperform the single-scan model in far range of over $12$ percentage point. Indeed, the semantic segmentation range on nuScenes reaches over $120$m. The addition of the postprocessing method,  $\text{prep (non-smearing)} \,+\, \text{model}_{m} \,+\, \text{postp}$ and $\text{prep (smearing)} \,+\, \text{model}_{m} \,+\, \text{postp}$, improves $\mIoU$ in the overall range of $1.6$ percentage point. 
In particular, it increases the performance in medium range from $60.8\%$ $\mIoU$ of $\text{model}_{s}$ to $67.0\%$ $\mIoU$ of $\text{prep (non-smearing)} \,+\, \text{model}_{m} \,+\, \text{postp}$ (over $6$ percentage point improvement), and it drastically improves in far range from $30.4\%$ $\mIoU$ of $\text{model}_{s}$ to $45.0\%$ $\mIoU$ of $\text{prep (smearing)} \,+\, \text{model}_{m} \,+\, \text{postp}$ (over $14$ percentage point improvement).
The improvement is particularly evident for example in class "pedestrian", where from $55.3\%$ $\mIoU$ in medium and $12.0\%$ $\mIoU$ in far ranges of $\text{model}_{s}$ it reaches respectively $71.6\%$ $\mIoU$ and $53.0\%$ $\mIoU$ in $\text{prep (smearing)} + \text{model}_{m} + \text{postp}$.
Still  the significant increments of performance are not well reflected in the overall $\mIoU$, due to the varying-density property of the points.

\begin{table}[hbt!]
\begin{center}
\caption{Range-based semantic segmentation results on nuScenes val set. The points are on average distributed as $72\%$ in close, $22\%$ in medium and $6\%$ in far ranges.}
\label{nuscenes_miou}
\scalebox{0.9}{
\begin{tabular}{l|l|lll}
\hline
Method & mIoU & close & medium & far
\\ \hline
$\text{model}_{s}$ & $77.8$ & $80.2$ & $60.8$ & $30.4$ \\
$\text{prep (non-smearing)} + \text{model}_{m}$ & $78.4$ & $79.9$ & $66.3$ & $41.4$ \\
$\text{prep (smearing)} + \text{model}_{m}$ & $78.1$ & $79.3$ & $66.2$ & $43.0$ \\
$\text{prep (non-smearing)} + \text{model}_{m} + \text{postp}$ & $\mathbf{79.4}$ & $\mathbf{81.0}$ & $\mathbf{67.0}$ & $43.1$ \\
$\text{prep (smearing)} + \text{model}_{m} + \text{postp}$ & $79.2$ & $80.6$ & $67.3$ & $\mathbf{45.0}$ 
\end{tabular}
}
\end{center}
\end{table}

\begin{table*}[hbt!]
\begin{center}
\captionsetup{justification=centering}
\caption{Semantic segmentation results on nuScenes val set overall and split on close, medium, far ranges.}
\label{nuscenes_classes}
\scalebox{0.65}{
\begin{tabular}{|l|l|l|llllllllllllllll|}
\cline{3-19}
\multicolumn{2}{c|}{} & mIoU  & \rotatebox[origin=c]{90}{barrier} & \rotatebox[origin=c]{90}{bicycle} & \rotatebox[origin=c]{90}{bus} & \rotatebox[origin=c]{90}{car} & \rotatebox[origin=c]{90}{ construction } & \rotatebox[origin=c]{90}{motorcycle} & \rotatebox[origin=c]{90}{pedestrian} & \rotatebox[origin=c]{90}{traffic cone} & \rotatebox[origin=c]{90}{trailer} & \rotatebox[origin=c]{90}{truck} & \rotatebox[origin=c]{90}{driveable} & \rotatebox[origin=c]{90}{other flat} & \rotatebox[origin=c]{90}{sidewalk} & \rotatebox[origin=c]{90}{terrain} & \rotatebox[origin=c]{90}{manmade} & \rotatebox[origin=c]{90}{vegetation}
\\ 
\hline
\multirow{5}{*}{\rotatebox{90}{overall}}
 & $\text{model}_{s}$ & $77.8$  & $77.3$ & $41.1$ & $\mathbf{95.0}$ & $91.2$ & $51.6$ & $85.7$ & $80.8$ & $64.5$ & $74.0$ & $85.0$ & $97.0$ & $71.8$ & $75.4$ & $74.3$ & $91.1$ & $89.3$ \\
 & $\text{prep (non-smearing)} + \text{model}_{m}$ & $78.4$  & $77.1$ & $44.9$ & $92.0$ & $\mathbf{93.5}$ & $57.2$ & $84.1$ & $81.4$ & $\mathbf{67.7}$ & $71.3$ & $85.6$ & $97.0$ & $69.5$ & $75.9$ & $\mathbf{75.4}$ & $91.8$ & $90.2$ \\
& $\text{prep (smearing)} + \text{model}_{m}$ & $78.1$  & $\mathbf{78.8}$ & $43.5$ & $91.3$ & $93.0$ & $\mathbf{65.1}$ & $80.0$ & $\mathbf{84.6}$ & $66.4$ & $61.2$ & $83.0$ & $97.0$ & $\mathbf{72.9}$ & $75.5$ & $75.2$ & $91.9$ & $90.0$ \\
& $\text{prep (non-smearing)} + \text{model}_{m} + \text{postp}$ & $\mathbf{79.4}$  & $78.0$ & $\mathbf{46.0}$ & $93.8$ & $91.8$ & $55.8$ & $\mathbf{88.0}$ & $82.8$ & $\mathbf{67.7}$ & $\mathbf{76.6}$ & $\mathbf{86.2}$ & $97.0$ & $72.8$ & $\mathbf{76.1}$ & $74.9$ & $92.0$ & $\mathbf{90.3}$ \\
& $\text{prep (smearing)} + \text{model}_{m} + \text{postp}$ & $79.2$  & $78.1$ & $45.1$ & $94.1$ & $91.5$ & $57.3$ & $87.4$ & $84.1$ & $67.6$ & $73.3$ & $85.5$ & $97.0$ & $72.8$ & $76.0$ & $74.9$ & $\mathbf{92.1}$ & $\mathbf{90.3}$ \\
\hline
\multirow{5}{*}{\rotatebox{90}{close}}
 & $\text{model}_{s}$ & $80.2$  & $79.3$ & $47.5$ & $97.7$ & $92.9$ & $53.1$ & $91.4$ & $\mathbf{87.5}$ & $69.0$ & $79.5$ & $88.4$ & $97.2$ & $72.9$ & $76.1$ & $74.6$ & $89.4$ & $86.5$ \\
 & $\text{prep (non-smearing)} + \text{model}_{m}$ & $79.9$  & $78.4$ & $49.4$ & $95.8$ & $\mathbf{95.1}$ & $57.2$ & $88.2$ & $85.7$ & $\mathbf{70.7}$ & $73.2$ & $88.5$ & $97.2$ & $70.3$ & $76.5$ & $\mathbf{75.4}$ & $89.5$ & $86.9$ \\
 & $\text{prep (smearing)} + \text{model}_{m}$ & $79.3$  & $\mathbf{80.1}$ & $47.0$ & $94.5$ & $94.7$ & $\mathbf{67.1}$ & $83.7$ & $88.2$ & $69.4$ & $59.5$ & $85.8$ & $97.2$ & $\mathbf{73.9}$ & $76.1$ & $75.3$ & $89.7$ & $86.7$ \\
 & $\text{prep (non-smearing)} + \text{model}_{m} + \text{postp}$ & $\mathbf{81.0}$  & $79.3$ & $\mathbf{50.7}$ & $\mathbf{97.8}$ & $93.2$ & $55.2$ & $\mathbf{92.2}$ & $87.1$ & $70.4$ & $\mathbf{81.6}$ & $\mathbf{89.0}$ & $97.2$ & $73.7$ & $\mathbf{76.6}$ & $74.8$ & $89.7$ & $86.9$ \\
 & $\text{prep (smearing)} + \text{model}_{m} + \text{postp}$ & $80.6$  & $79.4$ & $48.7$ & $97.4$ & $93.0$ & $56.0$ & $91.5$ & $87.4$ & $70.3$ & $78.7$ & $88.2$ & $97.2$ & $73.7$ & $76.5$ & $74.8$ & $\mathbf{89.8}$ & $\mathbf{87.0}$ \\
\hline
\multirow{5}{*}{\rotatebox{90}{medium}}
 & $\text{model}_{s}$ & $60.8$  & $55.4$ & $7.5$ & $\mathbf{77.4}$ & $74.4$ & $51.0$ & $28.0$ & $55.3$ & $28.3$ & $70.3$ & $61.8$ & $89.5$ & $48.7$ & $64.6$ & $72.3$ & $94.0$ & $93.4$ \\
 & $\text{prep (non-smearing)} + \text{model}_{m}$ & $66.3$  & $63.3$ & $26.0$ & $66.0$ & $76.2$ & $60.1$ & $46.1$ & $64.6$ & $43.8$ & $71.7$ & $65.6$ & $90.7$ & $53.6$ & $68.3$ & $75.4$ & $95.3$ & $94.6$ \\
 & $\text{prep (smearing)} + \text{model}_{m}$ & $66.2$  & $63.7$ & $26.9$ & $71.0$ & $75.2$ & $61.7$ & $40.8$ & $70.7$ & $43.2$ & $67.5$ & $63.0$ & $90.4$ & $52.4$ & $68.2$ & $74.7$ & $95.2$ & $94.6$ \\
 & $\text{prep (non-smearing)} + \text{model}_{m} + \text{postp}$ & $67.0$  & $63.7$ & $26.3$ & $65.5$ & $\mathbf{76.9}$ & $60.3$ & $\mathbf{48.3}$ & $65.1$ & $\mathbf{46.4}$ & $\mathbf{72.1}$ & $\mathbf{66.2}$ & $\mathbf{90.8}$ & $\mathbf{54.9}$ & $\mathbf{69.5}$ & $\mathbf{76.0}$ & $\mathbf{95.5}$ & $\mathbf{94.8}$ \\
 & $\text{prep (smearing)} + \text{model}_{m} + \text{postp}$ & $\mathbf{67.3}$  & $\mathbf{64.8}$ & $\mathbf{27.5}$ & $71.4$ & $75.4$ & $\mathbf{63.6}$ & $43.8$ & $\mathbf{71.6}$ & $45.6$ & $68.2$ & $64.7$ & $90.5$ & $54.5$ & $69.4$ & $75.5$ & $\mathbf{95.5}$ & $\mathbf{94.8}$ \\
\hline
\multirow{5}{*}{\rotatebox{90}{far}}
 & $\text{model}_{s}$ & $30.4$  & $23.6$ & $0.0$ & $29.3$ & $32.6$ & $24.8$ & $0.0$ & $12.0$ & $5.2$ & $32.3$ & $29.5$ & $24.7$ & $16.4$ & $8.9$ & $64.0$ & $91.4$ & $90.8$ \\
 & $\text{prep (non-smearing)} + \text{model}_{m}$ & $41.4$  & $38.3$ & $9.9$ & $29.7$ & $39.0$ & $43.1$ & $17.9$ & $36.8$ & $25.7$ & $42.8$ & $43.3$ & $33.8$ & $23.9$ & $18.3$ & $71.6$ & $94.0$ & $93.6$ \\
 & $\text{prep (smearing)} + \text{model}_{m}$ & $43.0$  & $39.8$ & $18.1$ & $30.7$ & $39.5$ & $47.5$ & $17.8$ & $50.1$ & $25.1$ & $43.2$ & $41.4$ & $31.4$ & $24.8$ & $19.3$ & $71.3$ & $94.1$ & $93.7$ \\
 & $\text{prep (non-smearing)} + \text{model}_{m} + \text{postp}$ & $43.1$  & $41.2$ & $9.2$ & $29.7$ & $40.6$ & $45.4$ & $\mathbf{23.6}$ & $38.9$ & $\mathbf{28.8}$ & $\mathbf{44.6}$ & $\mathbf{46.1}$ & $\mathbf{35.5}$ & $24.7$ & $19.4$ & $\mathbf{73.3}$ & $94.5$ & $94.1$ \\
 & $\text{prep (smearing)} + \text{model}_{m} + \text{postp}$ & $\mathbf{45.0}$  & $\mathbf{42.0}$ & $\mathbf{18.9}$ & $\mathbf{33.7}$ & $\mathbf{41.4}$ & $\mathbf{51.8}$ & $19.7$ & $\mathbf{53.0}$ & $28.4$ & $43.9$ & $44.9$ & $32.4$ & $\mathbf{27.6}$ & $\mathbf{20.2}$ & $73.1$ & $\mathbf{94.6}$ & $\mathbf{94.2}$ \\
\hline
\end{tabular}
}
\end{center}
\end{table*}

For both datasets, prep (smearing) turns out to be consistently superior in far range with respect to prep (non-smearing). This is because moving objects that are far to the ego-vehicle, even if they present dynamics not consistent with the egomotion, are still reasonable accumulated because the long distance makes less influential the velocity of those objects. Hence, in far range, the slightly smeared accumulation of moving objects is superior compared to the absence of accumulation. Overall, the performances of smearing vs non-smearing are comparable. This means that the possible noise introduced with the smearing, like in Fig. \ref{fig:smearing}, is overwhelmed by the transformer-based semantic segmentation model, as remarked in Subsec. \ref{subsec:SphereFormer}. Fig. \ref{fig:preds} shows an example of semantic segmentation improvement in far range given by our method, compared to the single-scan model.

\section{Conclusion} \label{sec:conclusion}

In this paper we presented our novel preprocessing and postprocessing method.
We demonstrated the effectiveness of the method through the validation of the generated multi-scan datasets on a SOTA voxel-based semantic segmentation model. In particular, we experimented with the generation of two multi-scan datasets by either removing or not points belonging to moving objects. From the numerical results, we verified that in both cases the performance achieved, in medium and far ranges, by our method significantly outperformed the single-scan SOTA semantic segmentation model.

In future work we hope to extend the preprocessing with smearing of moving objects by incorporating a temporal feature to each point. Indeed, in this work we just accumulated points coming from other timestamps and the semantic segmentation model treated them the same as reference points. However, by adding a temporal feature to each point, it would be possible to encode the timestamp the point belong to, via transformer-based architecture, as a temporal encoding. This could enhance the semantic segmentation performance and it would permit to classify objects as moving or static. In fact, a transformer-based architecture could in principle exploit voxel-wise temporal-positional encoding in order to rely more in voxels closer both in time and space and better filter out voxels corresponding to smearing of moving objects.
Finally, even if our method is designed to better suit downstream voxel-based architectures, the accumulation of voxels implicitly translates in an accumulation of points. For this reason, in future we want to extend our method for non-voxel-based architectures.

\bibliography{citation}
\bibliographystyle{unsrt}

\end{document}